\def\year{2019}\relax

\documentclass[letterpaper]{article} 
\usepackage{aaai19}  

\usepackage[T1]{fontenc}

\usepackage{tgtermes}
\usepackage{amsmath}
\usepackage{scalefnt,letltxmacro}
\LetLtxMacro{\oldtextsc}{\textsc}
\renewcommand{\textsc}[1]{\oldtextsc{\scalefont{1.10}#1}}
\usepackage[scaled=0.92]{PTSans}

\usepackage[usenames,dvipsnames]{xcolor}
\definecolor{shadecolor}{gray}{0.9}

\usepackage[english]{babel}
\usepackage[parfill]{parskip}
\usepackage{afterpage}
\usepackage{framed}
\usepackage{nicefrac}

\usepackage{lineno}

\usepackage{ragged2e}

\usepackage{graphicx}
\usepackage[labelfont=bf]{caption}
\usepackage{subfig}

\usepackage{booktabs}
\usepackage{tabu}

\usepackage{natbib}

\usepackage{listings}
\usepackage{fancyvrb}
\fvset{fontsize=\normalsize}

\usepackage[colorlinks,linktoc=all]{hyperref}
\usepackage[all]{hypcap}
\hypersetup{citecolor=Violet}
\hypersetup{linkcolor=black}
\hypersetup{urlcolor=MidnightBlue}

\usepackage[nameinlink]{cleveref}

\usepackage
[acronym,smallcaps,nowarn,section,nogroupskip,nonumberlist]{glossaries}
\glsdisablehyper{}
\crefname{appendix}{supplement}{}

\usepackage{listings}
\lstdefinestyle{alp_style}{
    commentstyle=\color{OliveGreen},
    numberstyle=\tiny\color{black!60},
    stringstyle=\color{BrickRed},
    basicstyle=\ttfamily\scriptsize,
    breakatwhitespace=false,
    breaklines=true,
    captionpos=b,
    keepspaces=true,
    numbers=none,
    numbersep=5pt,
    showspaces=false,
    showstringspaces=false,
    showtabs=false,
    tabsize=2
}
\usepackage{amsmath,amssymb,dsfont}
\usepackage{nicefrac}
\usepackage[usenames,dvipsnames]{xcolor}
\usepackage{adjustbox}
\usepackage{multirow}

\newcommand{\dd}{\mathrm{d}}
\newcommand{\RR}{\mathbb{R}}
\newcommand{\EE}{\mathbb{E}}

\newcommand{\uptoconst}{\mathrel{\overset{\makebox[0pt]{\mbox{\normalfont\small\sffamily c}}}{=}}}

\newcommand{\expec}[2]{\mathbb{E}_{#1}\left[#2\right]}
\newcommand{\boldf}{\boldsymbol{f}}
\newcommand{\boldy}{\boldsymbol{y}}
\newcommand{\bomega}{\boldsymbol{\omega}}
\newcommand{\bmu}{\boldsymbol{\mu}}
\newcommand{\boldeta}{\boldsymbol{\eta}}

\newcommand{\boldu}{\boldsymbol{u}}
\newcommand{\boldc}{\boldsymbol{c}}
\newcommand{\bkappa}{\boldsymbol{\kappa}}
\newcommand{\trace}{\text{tr}}
\newcommand{\Knm}{K_{nm}}
\newcommand{\invKmm}{K_{mm}^{-1}}
\newcommand{\Ktilde}{\widetilde{K}}

\begin{document}
%
\title{Efficient Gaussian Process Classification Using P\'olya-Gamma Data Augmentation}


\author{
Florian Wenzel,\textsuperscript{1,*}\;
Th\'eo Galy-Fajou,\textsuperscript{2,*}\;
Christan Donner,\textsuperscript{2}\;
Marius Kloft,\textsuperscript{1,3}\;
Manfred Opper\textsuperscript{2}\\
\textsuperscript{*}Contributed equally,\;
\textsuperscript{1}TU Kaiserslautern, Germany,\; \textsuperscript{2}TU Berlin, Germany,\; \textsuperscript{3}University of Southern California, USA\\
wenzelfl@hu-berlin.de, galy-fajou@tu-berlin.de, christian.donner@bccn-berlin.de,\\
kloft@cs.uni-kl.de, manfred.opper@tu-berlin.de
}


\maketitle
\begin{abstract}
We propose a scalable stochastic variational approach to GP classification building on P\'olya-Gamma data augmentation and inducing points. Unlike former approaches, we obtain closed-form updates based on natural gradients that lead to efficient optimization. We evaluate the algorithm on real-world datasets containing up to 11 million data points	and demonstrate that it is up to two orders of magnitude faster than the state-of-the-art while being competitive in terms of prediction performance.
\end{abstract}

\section{Introduction}
%
Gaussian processes (GPs) \cite{Rasmussen:2005:GPM:1162254} 
provide a popular  
Bayesian non-linear non-parametric method for regression and classification. 
Because of their ability of accurately adapting to data and thus achieving high prediction accuracy while providing well calibrated uncertainty estimates, GPs are a standard method in several application areas, including geospatial predictive modeling \cite{stein2012interpolation} and robotics \cite{dragiev2011gaussian}.

However, recent trends in data availability in the sciences and technology have made it necessary to develop algorithms capable of processing massive data \cite{john2014big}.
Currently, GP classification has limited applicability to big data. Naive inference typically scales cubic in the number of data points, and exact computation of posterior and marginal likelihood is intractable.

Nevertheless, the combination of so-called sparse Gaussian process techniques
with approximate inference methods, such as expectation propagation (EP)
or the variational approach, have enabled GP classification for datasets containing millions of data points
\cite{DBLP:conf/aistats/Hernandez-Lobato16,DBLP:conf/aistats/SalimbeniEH18}.

While these results are already impressive, we will show in this paper that
a speedup of up to two orders magnitudes can be achieved. 
Our approach is based on considering an augmented version of the original GP classification model
and
replacing the ordinary (stochastic) gradients for
optimization
by more efficient {\em natural gradients}, which is the standard Euclidean gradient multiplied by the inverse Fisher information matrix. Natural gradients
recently have been successfully used in 
a variety of variational inference problems \cite{conjugateVB,b-svm,sgdtm}.

Unfortunately, an efficient computation of the natural gradient
for the GP classification problem is not straight forward.
The use of the probit link function in
\citet{DBLP:conf/nips/DezfouliB15}; \citet{DBLP:conf/aistats/Hernandez-Lobato16}; \citet{probit}; \citet{DBLP:conf/aistats/SalimbeniEH18}
leads to expectations in the variational objective functions
that can only be computed by numerical quadrature, thus, preventing efficient optimization.

We derive a natural-gradient approach to variational inference in GP classification based on the {\em logit} link.
We exploit that the corresponding likelihood has an auxiliary variable representation as a continuous mixture of Gaussians involving P\'olya-Gamma random variables \cite{PG}.

Unlike former approaches,
our natural gradient updates can be computed in closed-form. Moreover, they have the advantage that they correspond to block-coordinate ascent updates and, therefore, learning rates close to one can be chosen.
This leads to a fast and stable algorithm which is simple to implement.
%
%
Our main contributions are as follows:
\begin{itemize}
	\item We present a Gaussian process classification model using a logit link function that is based on P\'olya-Gamma data augmentation
	and inducing points for Gaussian process inference.
	\item We derive an efficient inference algorithm based on stochastic variational inference and natural gradients. 
	All natural gradient updates are given in closed-form and do not rely on numerical quadrature methods or sampling approaches. Natural gradients have the
advantage that they
provide effective second-order optimization updates.
	\item In our experiments, we demonstrate that our approach drastically improves speed up to two orders of magnitude while being competitive in terms of prediction performance. We apply our method to massive real-world datasets up to 11 million points and demonstrate superior scalability. 
%
\end{itemize}%
The paper is organized as follows. In section \ref{sec:related_work} we discuss related work.
In section \ref{sec:model} we introduce our novel scalable GP classification model and in section \ref{sec:inference} we present an efficient variational inference
algorithm. Section \ref{sec:experiments} concludes with experiments. Our code is available via Github\footnote{\url{https://github.com/theogf/AugmentedGaussianProcesses.jl}}.


\section{Background and Related Work} \label{sec:related_work}

\paragraph{Gaussian process classification}
\citeauthor{Hensman2015} (\citeyear{Hensman2015}) consider Gaussian process classification with a probit inverse link function and suggest a variational Gaussian model that builds on inducing points.
By employing automatic differentiation, \citet{DBLP:conf/aistats/SalimbeniEH18} generalize this approach to use natural gradients in non-conjugate GP models.
\citet{DBLP:journals/corr/abs-1807-04489} consider natural gradient updates in the setting of variational inference with exponential families. 
Unlike our approach, these methods do not benefit from closed-form updates and have to resort to numerical approximations. 
Moreover, our approach has the advantage that a higher learning rate close to one can be chosen leading to updates that can be interpreted as block-coordinate ascent updates.

\citet{DBLP:conf/aistats/IzmailovNK18} use tensor train decomposition to allow for the training of GP models with billions of inducing points. The updates are not computed in closed-form and they do not use natural gradients.

\citet{DBLP:conf/nips/DezfouliB15} propose a general automated variational inference approach for sparse GP models with non-conjugate likelihood. Since they follow a black box approach and do not exploit model specific properties  they do not employ efficient optimization techniques.

\citet{DBLP:conf/aistats/Hernandez-Lobato16} follow an expectation propagation approach based on inducing points and have a similar computational cost as \citet{Hensman2015}.




\paragraph{P\'olya-Gamma data augmentation}
\citet{PG} introduced the idea of data augmentation in logistic models using the class of P\'olya-Gamma distributions. This allows for exact inference via Gibbs sampling or approximate variational inference schemes \cite{scott2013expectation}.

\citet{DBLP:conf/nips/LindermanJA15} extend this idea to multinomial models and discuss the application for Gaussian processes with multinomial observations but their approach does not scale to big datasets and they do not consider the concept of inducing points.



\section{Model} \label{sec:model}
The logit GP Classification model is defined as follows. 
Let $X = (\boldsymbol{x}_1,\ldots,\boldsymbol{x}_n) \in \RR^{d\times n}$ be the $d$-dimensional training points with labels $\boldsymbol{y}=(y_1,\ldots,y_n) \in \{-1,1\}^n$.
The likelihood of the labels is
\begin{align}\label{eq:likelihood}
p(\boldsymbol{y}\vert \boldsymbol{f}, X) = \prod_{i=1}^n\sigma(y_if(\boldsymbol{x}_i)),
\end{align}
where $\sigma(z)=(1+\exp(-z))^{-1}$ is the logit link function and $f$ is the latent decision function.
We place a GP prior over $f$ and obtain
the joint distribution of the labels and the latent GP
\begin{align}\label{eq:joint prob}
p(\boldsymbol{y},\boldsymbol{f}\vert X) = p(\boldsymbol{y}\vert \boldsymbol{f}, X)p(\boldsymbol{f}\vert X),
\end{align}
where $p(\boldsymbol{f}\vert X) = \mathcal{N}(\boldf | \boldsymbol{0},K_{nn})$ and $K_{nn}$ denotes the kernel matrix evaluated at the training points $X$. For the sake of clarity we omit the conditioning on $X$ in the following.

\subsection{P\'olya-Gamma data augmentation}
Due to the analytically inconvenient form of the likelihood function, inference for logit GP classification is a challenging problem.
We aim to remedy this issue by considering an augmented representation of the original model.
Later we will see that the augmented model is indeed advantageous as it leads to efficient closed-form updates in our variational inference scheme.

\citet{PG} introduced the class of P\'olya-Gamma random variables and proposed a data augmentation strategy for inference in models with binomial likelihoods. The augmented model has the appealing property that the likelihood of the latent function $\boldf$ is proportional to a Gaussian density when conditioned on the augmented P\'olya-Gamma variables. This allows for Gibbs sampling methods, where model parameters and P\'olya-Gamma variables can be sampled alternately from the posterior \cite{PG}. Alternatively, the augmentation scheme can be utilized to derive an efficient approximate inference algorithm in the variational inference framework, which will be pursued here. 



The P\'olya-Gamma distribution is defined as follows. The random variable $\omega \sim \mathrm{PG}(b, 0)$, $b > 0$ is defined by the moment generating function
\begin{align}
	\EE_{\mathrm{PG}(\omega\vert\, b,0)}[\exp(-\omega t)] = \frac{1}{\mathrm{cosh}^b(\sqrt{t/2})}.\label{eq:laplace_PG}
\end{align}
It can be shown that this is the Laplace transform of an 
infinite convolution of gamma distributions.
The definition is related to our problem by the fact that the logit link can be written in a form that involves the cosh function, namely $\sigma(z_i) = \exp(\frac{1}{2}z_i)  (2\cosh(\frac{z_i}{2}))^{-1}$.
In the following
we derive a representation of the logit link
in terms of P\'olya-Gamma variables.

First, we define the general $\mathrm{PG}(b, c)$ class which is derived by an exponential tilting of the $\mathrm{PG}(b, 0)$ density, it is given by
\begin{align*}
	\mathrm{PG}(\omega\vert\, b,c) \propto \exp(-\frac{c^2}{2}\omega) \mathrm{PG}(\omega\vert\, b,0).
\end{align*}
From the moment generating function \eqref{eq:laplace_PG} the first moment can be directly computed
\begin{align*}
\EE_{PG(\omega\vert b,c)}[\omega] = \frac{b}{2c}\tanh\left(\frac{c}{2}\right).
\end{align*}
For the subsequently presented variational algorithm these properties suffice and the full representation of the P\'olya-Gamma density $\mathrm{PG}(\omega\vert b, c)$ is not required.


We now adapt the data augmentation strategy based on P\'olya-Gamma variables for the GP classification model.
To do this we write the non-conjugate logistic likelihood function \eqref{eq:likelihood} in terms of P\'olya-Gamma variables  
\begin{align}
	\sigma(z_i) &= \left(1+\exp(-z_i)\right)^{-1}
	=  \frac{\exp(\frac{1}{2}z_i)}{2\cosh(\frac{z_i}{2})}\nonumber\\
	&= \frac{1}{2}\int \exp\left(\frac{z_i}{2} - \frac{z_i^2}{2}\omega_i\right)p(\omega_i)\dd\omega_i, \label{eq:logit_PG}
\end{align}
where $p(\omega_i) = \mathrm{PG}(\omega_i\vert 1,0)$ and by making use of \eqref{eq:laplace_PG}. 
For more details see \citet{PG}.
Using this identity and substituting $z_i = y_if(x_i)$ we augment the joint density \eqref{eq:joint prob} with P\'olya-Gamma variables
\begin{align}
p(\boldsymbol{y},\boldsymbol{\omega},\boldf) 
\propto \exp\left(\frac{1}{2}\boldy^\top\boldf - \frac{1}{2}\boldf^\top \Omega\boldf\right)p(\boldf)p(\bomega), \label{eq:pg_joint}
\end{align}
where $\Omega = \text{diag} (\bomega)$ is the diagonal matrix of the P\'olya-Gamma variables $\{ \omega_i \}$. In contrast to the original model~\eqref{eq:joint prob} the augmented model is conditionally conjugate forming the basis for deriving closed-form updates in section~\ref{sec:inference}.

Interestingly, employing a structured mean-field variational inference approach (cf. section~\ref{sec:inference})
to the plain P\'olya-Gamma augmented model~\eqref{eq:pg_joint} leads to the same bound for GP classification derived by \citet{Gibbs_MacKay97b}.
This is an interesting new perspective on this bound since they do not employ a data augmentation approach. We provide a proof in appendix~\ref{sec:appendix_mackay}.
Our approach goes beyond \citet{Gibbs_MacKay97b} by providing a fully Bayesian perspective, including a sparse GP prior (section \ref{sec:sparse_GP}) in the model and proposing a scalable inference algorithm based on natural gradients (section \ref{sec:inference}).


\subsection{Sparse Gaussian process}
\label{sec:sparse_GP}
Inference in GP models typically has the computational complexity $\mathcal{O}(n^3)$.
We obtain a scalable approximation of our model and focus on inducing point methods \cite{NIPS2005_sparseGP}.
We follow a similar approach as in \citet{Hensman2015} and reduce the complexity to $\mathcal{O}(m^3)$, where $m$ is number of inducing points.

We augment the latent GP $f$ with $m$ additional input-output pairs $(Z_1,u_1),\ldots,(Z_m,u_m)$, termed as \emph{inducing inputs} and \emph{inducing variables}.
The function values of the GP $\boldf$ and the inducing variables $\boldu = (u_1,\ldots,u_m)$ are connected via
\begin{align}
\begin{split} \label{eq:inducing_gp}
p(\boldf|\boldu) &= \mathcal{N}\left(\boldf|K_{nm}K_{mm}^{-1}\boldu, \widetilde{K}\right)\\
p(\boldu) &= \mathcal{N}\left(\boldu|0,K_{mm}\right),
\end{split}
\end{align}
where $K_{mm}$ is the kernel matrix resulting from evaluating the kernel function between all inducing inputs, $K_{nm}$ is the cross-kernel matrix between inducing inputs and training points and $\widetilde{K} = K_{nn} - K_{nm}K^{-1}_{mm}K_{mn}$.
Including the inducing points in our model gives the augmented joint distribution
\begin{align}
p(\boldy,\bomega,\boldf,\boldu) &= p(\boldy|\bomega,\boldf)p(\bomega)p(\boldf|\boldu)p(\boldu)
\end{align}
Note that the original model \eqref{eq:joint prob} can be recovered by marginalizing $\bomega$ and $\boldu$.


\section{Inference} \label{sec:inference}
The goal of Bayesian inference is to compute the posterior of the latent model variables. Because this problem is intractable for the model at hand, we employ variational inference to map the inference problem to a feasible optimization problem.
We first chose a family of tractable variational distributions and select the best candidate by minimizing the Kullback-Leibler divergence between the variational distribution and the posterior.
This is equivalent to optimizing a lower bound on the marginal likelihood, known as evidence lower bound (ELBO) \cite{Jordan:1999:IVM:339248.339252,Wainwright:2008:GME:1498840.1498841}.

In the following we develop a stochastic variational inference (SVI) algorithm that enables stochastic optimization based on natural gradient updates which are given in closed-form.

\subsection{Why use natural gradients?}
Using the natural gradient over the standard Euclidean gradient is favorable since
natural gradients are invariant to reparameterization of the variational family \cite{infogeom,naturalgrad} and provide effective second-order optimization updates \cite{amari98natural,JMLR:v14:hoffman13a}. 


The superiority of using natural gradients in our approach can be explained by the following. We reformulate the GP classification model as an augmented model which is conditionally conjugate. 
When using a learning rate of one, the natural gradient updates correspond to block-coordinate ascent updates, i.e. in each iteration each parameter is set to its optimal value given the remaining parameters (see appendix \ref{sec:appendix_natural_grad} and \citet{JMLR:v14:hoffman13a}). In practice, we employ stochastic variational inference, i.e. we only use mini-batches of the data to obtain a noisy version of the natural gradient. In this setting, learning rates slightly less than one have to be chosen.

This is in contrast to former natural gradient based approaches, e.g. \cite{DBLP:conf/aistats/SalimbeniEH18}, that focus on the original non-conjugate GP classification model. Although they benefit from using natural gradients, they have the disadvantage that their updates do not correspond to coordinate-ascent updates. Thus, learning rates that are much smaller that one have to be used to assure convergence.

Therefore, in our approach, we can use much higher learning rates and optimization is faster and more stable which we demonstrate in the experiments.


\subsection{Variational approximation}
\label{sec:var_approximation}

We aim to approximate the posterior of the inducing points $p(\boldu | \boldy)$ and 
apply the methodology of variational inference to the marginal joint distribution $p(y,\omega, u) = p(\boldy | \bomega, \boldu) p(\bomega) p(\boldu)$. Following a similar approach as \citet{Hensman2015}, we apply Jensen's inequality to obtain a tractable lower bound on the log-likelihood of the labels
\begin{align}
	\log p(\boldy | \bomega, \boldu) &= \log \EE_{p(f|u)}[p(\boldy | \bomega, f)] \nonumber\\
		&\ge \EE_{p(f|u)}[\log p(\boldy | \bomega, f)]. \label{eq:ineq_likelihood}
\end{align}
By this inequality we construct a variational lower bound on the evidence
\begin{align*}
	\log p(\boldy)
	&\ge \EE_{q(\boldu, \bomega)}[ \log p(\boldy | \boldu, \bomega)]
	- \mathrm{KL}\left( q(\boldu, \bomega) || p(\boldu, \bomega) \right)\\
		&\ge \EE_{p(\boldf|\boldu) q(\boldu) q(\bomega)}[\log p(\boldy | \bomega, \boldf)]\\
	    &\quad -\mathrm{KL}\left( q(\boldu, \bomega) || p(\boldu, \bomega) \right)\\
		&=: \mathcal{L},
\end{align*}
where the first inequality is the usual evidence lower bound (ELBO) in variational inference and the second inequality is due to \eqref{eq:ineq_likelihood}. 

We follow a structured mean-field approach \cite{Wainwright:2008:GME:1498840.1498841} and assume independence between the inducing variables $u$ and P\'olya-Gamma variables $\omega$, yielding a variational distribution of the form $q(u,\omega) = q(u) q(\omega)$. Setting the functional derivative of $\mathcal{L}$ w.r.t. $q(u)$ and $q(\omega)$ to zero, respectively, results in the following consistency condition for the maximum,
\begin{align}
q(\boldu,\bomega)=q(\boldu)\prod_i q(\omega_i), \label{eq:var_family}
\end{align}
with $q(\omega_i)=\mathrm{PG}(\omega_i|1,c_i)$ and $q(\boldu)=\mathcal{N}(\boldu|\bmu,\Sigma)$. Remarkably, we do not have to use the full P\'olya-Gamma class $\mathrm{PG}(\omega_i|b_i,c_i)$, but instead consider the restricted class $b_i=1$ since it already contains the optimal distribution.

We use \eqref{eq:var_family} as variational family which is parameterized by the variational parameters $\{\bmu, \Sigma, \boldc\}$ and
obtain a closed-form expression of the variational bound
\begin{align}
&\mathcal{L}(\boldc, \bmu, \Sigma) \nonumber\\
&= \EE_{p(\boldf|\boldu) q(\boldu) q(\bomega)}[\log p(\boldy | \bomega, \boldf)]
- \mathrm{KL}\left( q(\boldu, \bomega) || p(\boldu, \bomega) \right) \nonumber\\
&\uptoconst \frac{1}{2}\bigg(\log|\Sigma|-\log|K_{mm}|)- \trace(\invKmm\Sigma)
-\bmu^\top\invKmm\bmu\nonumber\\
&\quad +\sum_i \Big\{y_i\bkappa_i\bmu -
\theta_i\left(\Ktilde_{ii}-\bkappa_i\Sigma\bkappa_i^\top 
 -\bmu^\top\bkappa_i^\top\bkappa_i\bmu\right)\nonumber\\
&\quad + c_i^2 \theta_i - 2\log \text{cosh}\frac{c_i}{2}\Big\}\bigg),	\label{eq:variational_bound}
\end{align}
where $\theta_i = \frac{1}{2c_i}\tanh\left(\frac{c_i}{2}\right)$ and $\bkappa_i = K_{im}\invKmm$. 
Remarkably, all intractable terms involving expectations of $\log \mathrm{PG}(\omega_i\vert 1, 0)$ cancel out.
Details are provided in appendix \ref{sec:appendix_variational_bound}.



\subsection{Stochastic variational inference}



Our algorithm alternates between updates of the local variational parameters $\boldc$ and global parameters $\bmu$ and $\Sigma$. In each iteration we update the para\-meters based on a mini-batch of the data $\mathcal{S}\subset \{1,...,n\}$ of size $s = |\mathcal{S}|$.

We update the \emph{local parameters} $\boldc_\mathcal{S}$ in the mini-batch $\mathcal{S}$ by employing coordinate ascent. To this end, we fix the global parameters and analytically compute the unique maximum of \eqref{eq:variational_bound} w.r.t. the local parameters, leading to the updates
\begin{align}
\begin{split}
c_i &= \sqrt{\Ktilde_{ii}+\bkappa_i\Sigma\bkappa_i^\top + \bmu^\top\bkappa_i^\top\bkappa_i\bmu}
\end{split} \label{eq:local_updates}
\end{align}
for $i \in \mathcal{S}.$

We update the \emph{global parameters} by employing stochastic optimization of the variational bound \eqref{eq:variational_bound}.
The optimization is based on stochastic estimates of the natural gradients of the global parameters.
We use the natural parameterization of the variational Gaussian distribution, i.e., the parameters $\boldeta_1:=\Sigma^{-1} \bmu$ and $\eta_2 = -\frac{1}{2}\Sigma^{-1}$. Using the natural parameters results in simpler and more effective updates.
The natural gradients based on the mini-batch $\mathcal{S}$ are given by
\begin{align}
\begin{split} \label{eq:natural_gradients}
\widetilde \nabla_{\boldeta_1} \mathcal{L}_{\mathcal{S}} &= \frac{n}{2s}\bkappa_{\mathcal{S}}^\top\boldy_{\mathcal{S}} - \boldeta_1\\
\widetilde \nabla_{\eta_2} \mathcal{L}_{\mathcal{S}} &= -\frac{1}{2}\left(\invKmm+\frac{n}{s}\bkappa_{\mathcal{S}}^\top\Theta_{\mathcal{S}}\bkappa_{\mathcal{S}}\right) - \eta_2,
\end{split}
\end{align}
where $\Theta = \text{diag}(\boldsymbol{\theta})$ and $\theta_i = \frac{1}{2c_i}\tanh\left(\frac{c_i}{2}\right)$.
The factor $\frac{n}{s}$ is due to the rescaling of the mini-batches.
The global parameters are updated according to a stochastic natural gradient ascent scheme. We employ the adaptive learning rate method described by \citet{adaptiveSVI}.


The natural gradient updates always lead to a positive definite covariance matrix\footnote{This follows directly since $K_{mm}$ and $\Theta$ are positive definite.}
and in contrast to \citet{Hensman2015} our implementation does not require any assurance for positive-definiteness of the variational covariance matrix $\Sigma$. 
Details for the derivation of the updates can be found in appendix \ref{sec:appendix_updates}.
The complexity of each iteration in the inference scheme is $\mathcal{O}(m^3)$, due to the inversion of the matrix $\eta_2$.

\paragraph{On the quality of the approximation}
In other applications of variational inference to GP classification, 
one tries to approximate the posterior directly by a Gaussian $q^*(f)$
which minimizes the Kullback-Leibler divergence between the variational distribution and the true posterior \cite{Hensman2015}. 
On the other hand, in our paper, we apply variational inference 
to the augmented model, looking for the best distribution that factorizes
in the P\'olya-Gamma variables $\omega_i$ and the original function $f$. This approach also yields a Gaussian approximation $q(f)$ as a factor in the optimal density. 
Of course
$q(f)$ will be different from the ‘optimal’ $q^*(f)$. We could however argue that
asymptotically, in the limit of a large number of data, the predictions given by both
densities may not be too different, as the posterior
uncertainty for both densities should become small \cite{Opper:2009:VGA:1512583.1512590}. 

It would be interesting to see how the ELBOs of the two variational approaches,
which both give a lower bound on the likelihood of the data,
differ. Unfortunately, such a computation would require the knowledge of the optimal $q^*(f)$.
However, we can obtain some estimate of this difference when we assume that 
we use the {\em same} Gaussian density $q(f)$ for both bounds as an approximation.
In this case, we obtain 
\begin{align*}
  \mathcal{L}_\text{orig} - \mathcal{L}_\text{augmented}
  &= \EE_{q(f)}[\text{KL}\left(q(\omega) || p(\omega|f,y)\right)].
\end{align*}
This lower bound on the gap is small if on average the variational approximation $q(\omega)$ is close to the posterior $p(\omega|f,y)$. 
For the sake of simplicity we consider here the non-sparse case, i.e. the inducing points equal the training points ($f=u$). However, it is straight-forward to extend the results also to the sparse case.

We empirically investigate the quality of our approximation in experiment~\ref{sec:exp_quality}.

\paragraph{Predictions}
The approximate posterior of the GP values and inducing variables is given by $q(\boldf ,\boldu) = p(\boldf |\boldu)q(\boldu)$, where $q(\boldu) = \mathcal{N}(\boldu | \bmu, \Sigma)$ denotes the optimal variational distribution. To predict the latent function values $f_*$ at a test point $x_*$ we substitute our approximate posterior into the standard predictive distribution
\begin{align}
	&p(f_* | y) = \int p(f_* | \boldf , \boldu) p(\boldf ,\boldu | y) \dd \boldf \dd \boldu \nonumber\\
		&\approx \int p(f_* | \boldf , \boldu) p(\boldf | \boldu) q(\boldu) \dd \boldf \dd \boldu \nonumber\\
		&= \int p(f_* | \boldu) q(\boldu) \dd \boldu 
		=  {\cal N}\left(f^* | \mu_*, \; \sigma^2_*\right),
		\label{eq:predictive_distr}
\end{align}
where the prediction mean is $\mu_*=K_{*m}K_{mm}^{-1}\bmu$ and the variance $\sigma^2_*=K_{**} + K_{*m}K_{mm}^{-1}(\Sigma K_{mm}^{-1} - I)K_{m*}$. The matrix $K_{*m}$ denotes the kernel matrix between the test point and the inducing points and $K_{**}$ the kernel value of the test point.
The distribution of the test labels is easily computed by applying the logit link function to \eqref{eq:predictive_distr},
\begin{align}
	p(y_*=1|y) = \int \sigma(f_*) p(f_* | y)  \dd f_*. \label{eq:test_distribution}
\end{align}
This integral is analytically intractable but can be computed numerically by quadrature methods. This is adequate and fast since the integral is only one-dimensional.

Computing the mean and the variance of the predictive distribution has complexity $\mathcal{O}(m)$
and $\mathcal{O}(m^2)$, respectively.




\paragraph{Optimization of the hyperparameters}
We select the optimal kernel hyperparameters by maximizing the marginal likelihood $p(y|h)$, where $h$ denotes the set of hyperparameters
(this approach is called empirical Bayes \cite{EB89}).
We follow an approximate approach and optimize the fitted variational lower bound ${\cal L}(h)$ \eqref{eq:variational_bound} as a function of $h$ by alternating between optimization steps w.r.t. the variational parameters and the hyperparameters \cite{cSGD}.






\section{Experiments} \label{sec:experiments}

We compare our proposed method, efficient Gaussian process classification (\textsc{x-gpc}), with the state-of-the-art methods \textsc{svgpc} \cite{DBLP:conf/aistats/SalimbeniEH18}, provided in the package GPflow\footnote{We use GPflow version 1.2.0.} \cite{GPflow2017}, which builds on TensorFlow
and the EP approach \textsc{epgpc} by \citet{DBLP:conf/aistats/Hernandez-Lobato16}, implemented in R. All methods are applied to real-world datasets containing up to 11 million data points.


In all experiments a squared exponential covariance function with a common length scale parameter for each dimension, an amplitude parameter and an additive noise parameter is used. The kernel hyperparameters are initialized to the same values and optimized using Adam \cite{journals/corr/KingmaB14}, while inducing points location are initialized via k-means++ \cite{arthur2007k} and kept fixed during training.
The SVI based methods, \textsc{x-gpc} and \textsc{svgpc}, use an adaptive learning rate.
All algorithms are run on a single CPU.
We experiment on 12 datasets from the OpenML website and the UCI repository ranging from 768 to 11 million data points. 
%
In the first experiment (section \ref{sec:exp_quality}), we examine the quality of the approximation provided by \textsc{x-gpc}. In the next experiment, we evaluate the prediction performance and run time of \textsc{x-gpc} and \textsc{svgpc} and \textsc{epgpc} on several real-world datasets.
Finally, in \ref{sec:exp_inducing}, we examine the sensitivity of all methods to the number of inducing points.

\subsection{Quality of the approximation}
\label{sec:exp_quality}
\begin{figure}
\centering
\includegraphics[width=0.85\columnwidth]{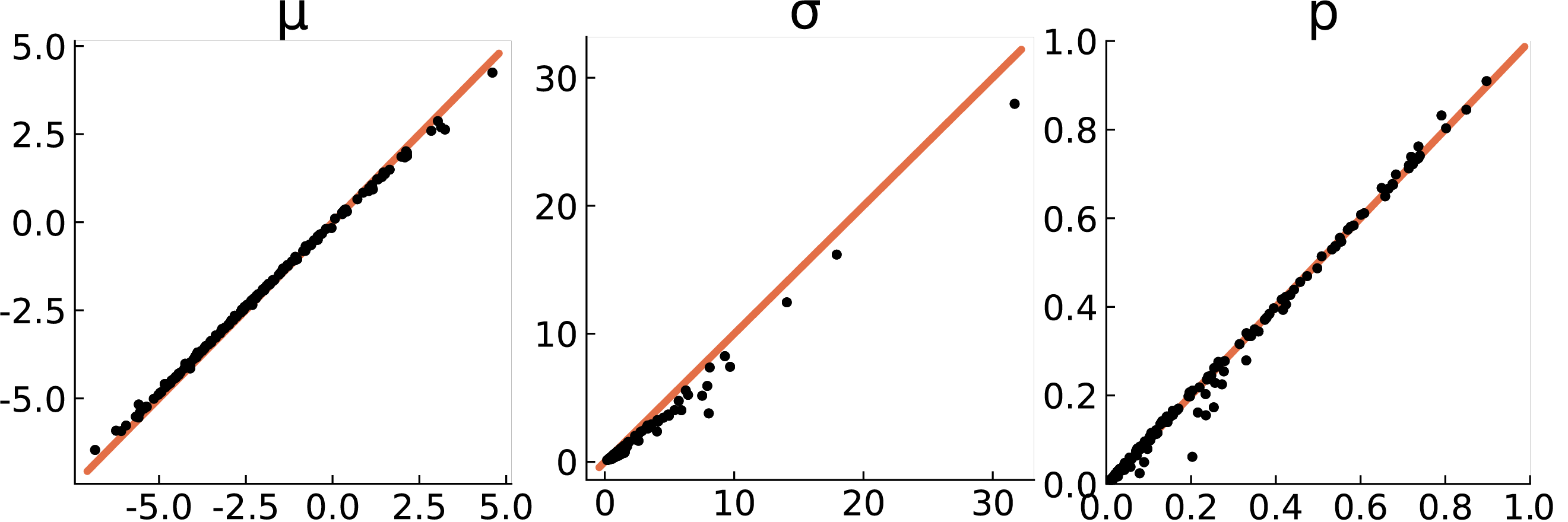}
\caption{Posterior mean ($\mu$), variance ($\sigma$) and predictive marginals ($p$) of the Diabetes dataset. 
Each plot shows the MCMC ground truth on the x-axis and the estimated value of our model on the y-axis. Our approximation is very close to the ground truth.}
\label{fig:qualiy_of_approx}
\end{figure}

We empirically examine the quality of the variational approximation provided by our method.
In Fig.~\ref{fig:qualiy_of_approx}, we compare the approximations to the true posterior obtained by employing an asymptotically correct Gibbs sampler \cite{Polson11,DBLP:conf/nips/LindermanJA15}. We compare the posterior mean and variance as well as the prediction probabilities with the ground truth.  Since the Gibbs sampler does not scale to large datasets we experiment on the small Diabetes dataset. In Fig.~\ref{fig:qualiy_of_approx} we plot the approximated values vs. the ground truth. 
We find that our approximation is very close to the true posterior.

\subsection{Numerical comparison}
\label{sec:exp_performance}

\begin{table}[h!]\centering
\begin{adjustbox}{max width=\columnwidth}
\footnotesize
\begin{tabular}{|l|l|l|l|l|}
Dataset & & \textbf{X-GPC} & SVGPC & EPGPC \\\hline
\multirow{1}{*}{aXa} & Error & $ \mathbf{0.17 \pm 0.07} $ & $ \mathbf{0.17 \pm 0.07} $ & $ \mathbf{ 0.17 \pm 0.07 } $\\
$ n=36,974 $ & NLL & $ \mathbf{ 0.29 \pm 0.13 } $ & $ 0.36 \pm 0.13 $ & $ 0.34 \pm 0.13 $\\
$ d=123 $ & Time & $ \mathbf{ 47 \pm 2.2 } $ & $ 451 \pm 7.8 $ & $ 214 \pm 4.8 $\\\hline
\multirow{1}{*}{Bank Market.} & Error & $ 0.14 \pm 0.12 $ & $ \mathbf{0.12 \pm 0.12} $ & $ \mathbf{ 0.12 \pm 0.13 } $\\
$ n=45,211 $ & NLL & $ \mathbf{ 0.27 \pm 0.22 } $ & $ 0.31 \pm 0.26 $ & $ 0.33 \pm 0.20 $\\
$ d=43 $ & Time & $ \mathbf{ 9 \pm 1.5 } $ & $ 205 \pm 6.6 $ & $ 46 \pm 3.5 $\\\hline
\multirow{1}{*}{Click Pred.} & Error & $ \mathbf{ 0.17 \pm 0.00 } $ & $ \mathbf{0.17 \pm 0.00} $ & $ \mathbf{0.17 \pm 0.01} $\\
$ n=399,482 $ & NLL & $ \mathbf{ 0.39 \pm 0.07 } $ & $ 0.46 \pm 0.00 $ & $ 0.46 \pm 0.01 $\\
$ d=12 $ & Time & $ \mathbf{4.5 \pm 1.3} $ & $ 102 \pm 3.0 $ & $ 8.1 \pm 0.45 $\\\hline
\multirow{1}{*}{Cod RNA} & Error & $ \mathbf{ 0.04 \pm 0.00 } $ & $ \mathbf{0.04 \pm 0.00} $ & $ \mathbf{0.04 \pm 0.00} $\\
$ n=343,564 $ & NLL & $ \mathbf{ 0.11 \pm 0.03 } $ & $ 0.13 \pm 0.00 $ & $ 0.12 \pm 0.00 $\\
$ d=8 $ & Time & $ \mathbf{ 3.7 \pm 0.13 } $ & $ 115 \pm 4.3 $ & $ 869 \pm 5.2 $\\\hline
\multirow{1}{*}{Diabetes} & Error & $ \mathbf{0.23 \pm 0.07} $ & $ \mathbf{ 0.23 \pm 0.06 } $ & $ 0.24 \pm 0.06 $\\
$ n=768 $ & NLL & $ \mathbf{ 0.47 \pm 0.11 } $ & $ 0.47 \pm 0.10 $ & $ 0.48 \pm 0.09 $\\
$ d=8 $ & Time & $ 8.8 \pm 0.12 $ & $ 150 \pm 5.1 $ & $ \mathbf{ 8 \pm 0.45 } $\\\hline
\multirow{1}{*}{Electricity} & Error & $ \mathbf{ 0.24 \pm 0.06 } $ & $ 0.26 \pm 0.06 $ & $ 0.26 \pm 0.06 $\\
$ n=45,312 $ & NLL & $ \mathbf{ 0.31 \pm 0.17 } $ & $ 0.53 \pm 0.08 $ & $ 0.53 \pm 0.06 $\\
$ d=8 $ & Time & $ \mathbf{8.2 \pm 0.48} $ & $ 356 \pm 6.9 $ & $ 13.5 \pm 1.50 $\\\hline
\multirow{1}{*}{German} & Error & $ \mathbf{0.25 \pm 0.12} $ & $ \mathbf{ 0.25 \pm 0.11 } $ & $ 0.26 \pm 0.13 $\\
$ n=1,000 $ & NLL & $ \mathbf{ 0.44 \pm 0.17 } $ & $ 0.51 \pm 0.15 $ & $ 0.53 \pm 0.11 $\\
$ d=20 $ & Time & $ 17 \pm 0.42 $ & $ 374 \pm 7.3 $ & $ \mathbf{ 5.2 \pm 0.03 } $\\\hline
\multirow{1}{*}{Higgs} & Error & $ \mathbf{ 0.33 \pm 0.01 } $ & $ 0.45 \pm 0.01 $ & $ 0.38 \pm 0.01 $\\
$ n=11,000,000 $ & NLL & $ \mathbf{ 0.55 \pm 0.13 } $ & $ 0.69 \pm 0.00 $ & $ 0.66 \pm 0.00 $\\
$ d=28 $ & Time & $ \mathbf{ 23 \pm 0.88 } $ & $ 294 \pm 54 $ & $ 8732 \pm 867 $\\\hline
\multirow{1}{*}{IJCNN} & Error & $ 0.03 \pm 0.01 $ & $ 0.06 \pm 0.01 $ & $\mathbf{ 0.02 \pm 0.01} $\\
$ n=141,691 $ & NLL & $ 0.10 \pm 0.03 $ & $ 0.15 \pm 0.07 $ & $ \mathbf{0.09 \pm 0.04} $\\
$ d=22 $ & Time & $ \mathbf{17 \pm 0.44} $ & $ 1033 \pm 45 $ & $ 756 \pm 8.6 $\\\hline
\multirow{1}{*}{Mnist} & Error & $  0.14 \pm 0.01  $ & $ 0.44 \pm 0.13 $ & $ \mathbf{0.12\pm 0.01} $\\
$ n=70,000 $ & NLL & $ \mathbf{ 0.24 \pm 0.10 } $ & $ 0.66 \pm 0.11 $ & $ 0.27 \pm 0.01 $\\
$ d=780 $ & Time & $ \mathbf{200 \pm 5.5} $ & $ 991 \pm 23 $ & $ 806 \pm 5.2 $\\\hline
\multirow{1}{*}{Shuttle} & Error & $ \mathbf{ 0.01 \pm 0.01 } $ & $ \mathbf{0.01 \pm 0.00} $ & $ \mathbf{0.01 \pm 0.01} $\\
$ n=58,000 $ & NLL & $ \mathbf{ 0.07 \pm 0.01 } $ & $ \mathbf{0.07 \pm 0.00} $ & $\mathbf{ 0.07 \pm 0.01 }$\\
$ d=9 $ & Time & $ \mathbf{ 0.01 \pm 0.00 } $ & $ 7.5 \pm 0.7 $ & $ 100 \pm 0.63 $\\\hline
\multirow{1}{*}{SUSY} & Error & $ \mathbf{ 0.21 \pm 0.00 } $ & $ 0.22 \pm 0.00 $ & $ 0.22 \pm 0.00 $\\
$ n=5,000,000 $ & NLL & $ \mathbf{ 0.31 \pm 0.10 } $ & $ 0.49 \pm 0.01 $ & $ 0.50 \pm 0.00 $\\
$ d=18 $ & Time & $ \mathbf{ 14 \pm 0.29 } $ & $ 10,000 $ & $ 10,000 $\\\hline
\multirow{1}{*}{wXa} & Error & $ \mathbf{0.03 \pm 0.01} $ & $ 0.04 \pm 0.01 $ & $ \mathbf{ 0.03 \pm 0.01 } $\\
$ n=34,780 $ & NLL & $ 0.27 \pm 0.07 $ & $ 0.25 \pm 0.07 $ & $ \mathbf{ 0.19 \pm 0.06 } $\\
$ d=300 $ & Time & $ 66 \pm 16 $ & $ 612 \pm 11 $ & $ \mathbf{ 1.4 \pm 0.10 } $\\\hline
\end{tabular}
\end{adjustbox}
\caption{Average test prediction error, negative test log-likelihood (NLL) and time in seconds along with one standard deviation. Best values are highlighted.
}
\label{tab:performance}
\end{table}

\begin{figure*}
\centering
\includegraphics[width=0.66\textwidth]{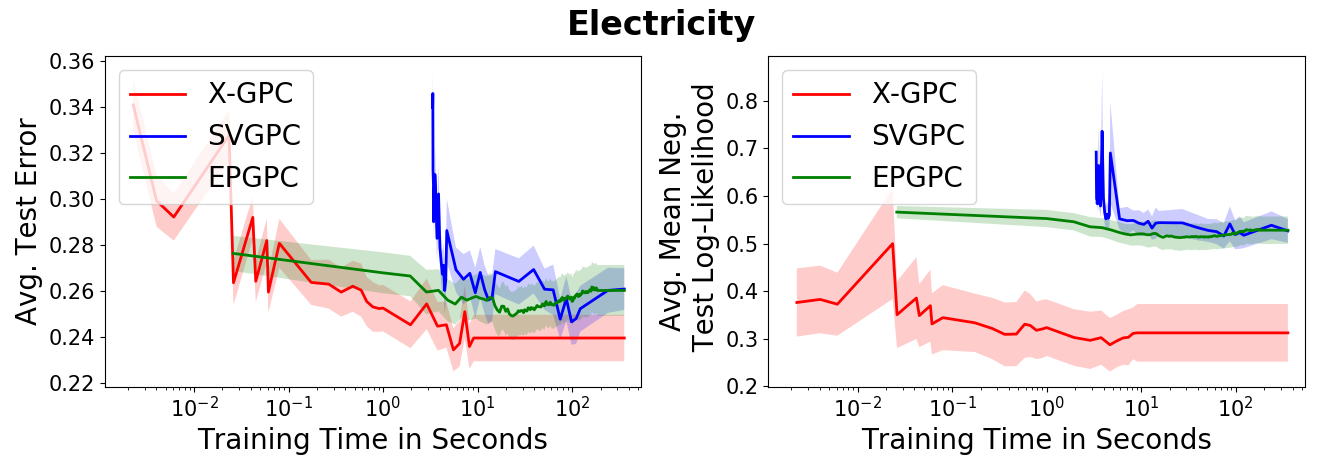}
\includegraphics[width=0.66\textwidth]{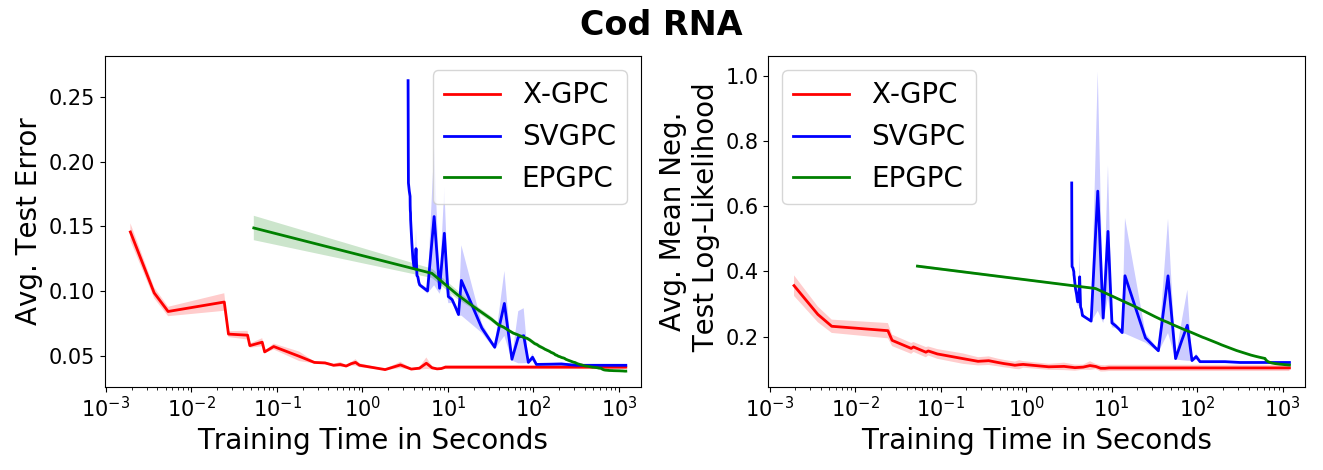}
\includegraphics[width=0.66\textwidth]{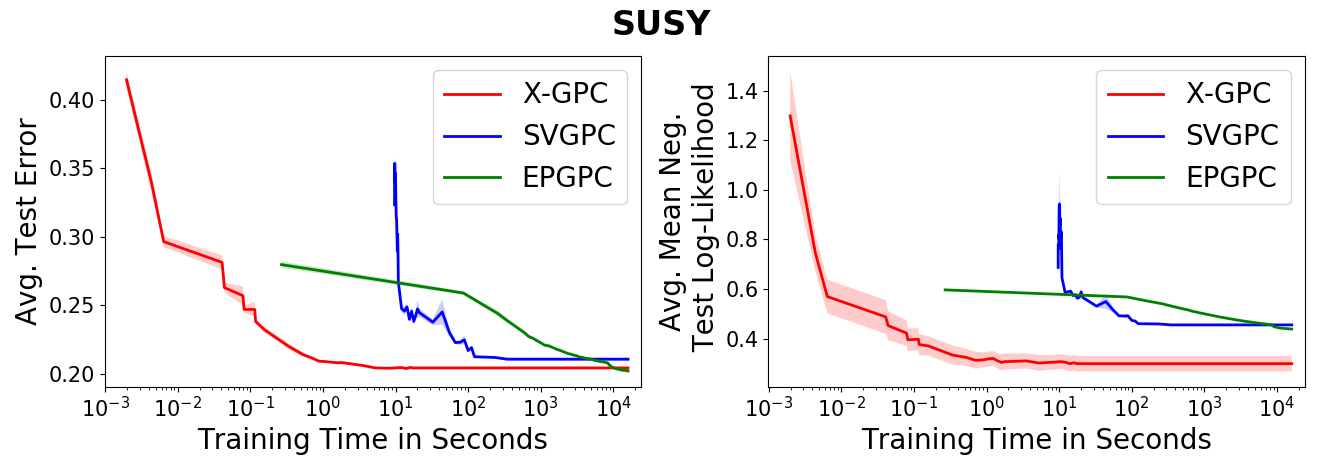}
\caption{Average negative test log-likelihood and average test prediction error as a function of training time (seconds in a $\log_{10}$ scale) on the datasets Electricity (45,312 points), Cod RNA (343,564 points) and SUSY (5 million points). \textsc{x-gpc} (proposed) reaches values close to the optimum after only a few iterations, whereas \textsc{svgpc} and \textsc{epgpc} are one to two orders of magnitude slower.} 
\label{fig:timeexperiment}
\end{figure*}%

\begin{figure*}
\centering
\includegraphics[width=0.7\textwidth]{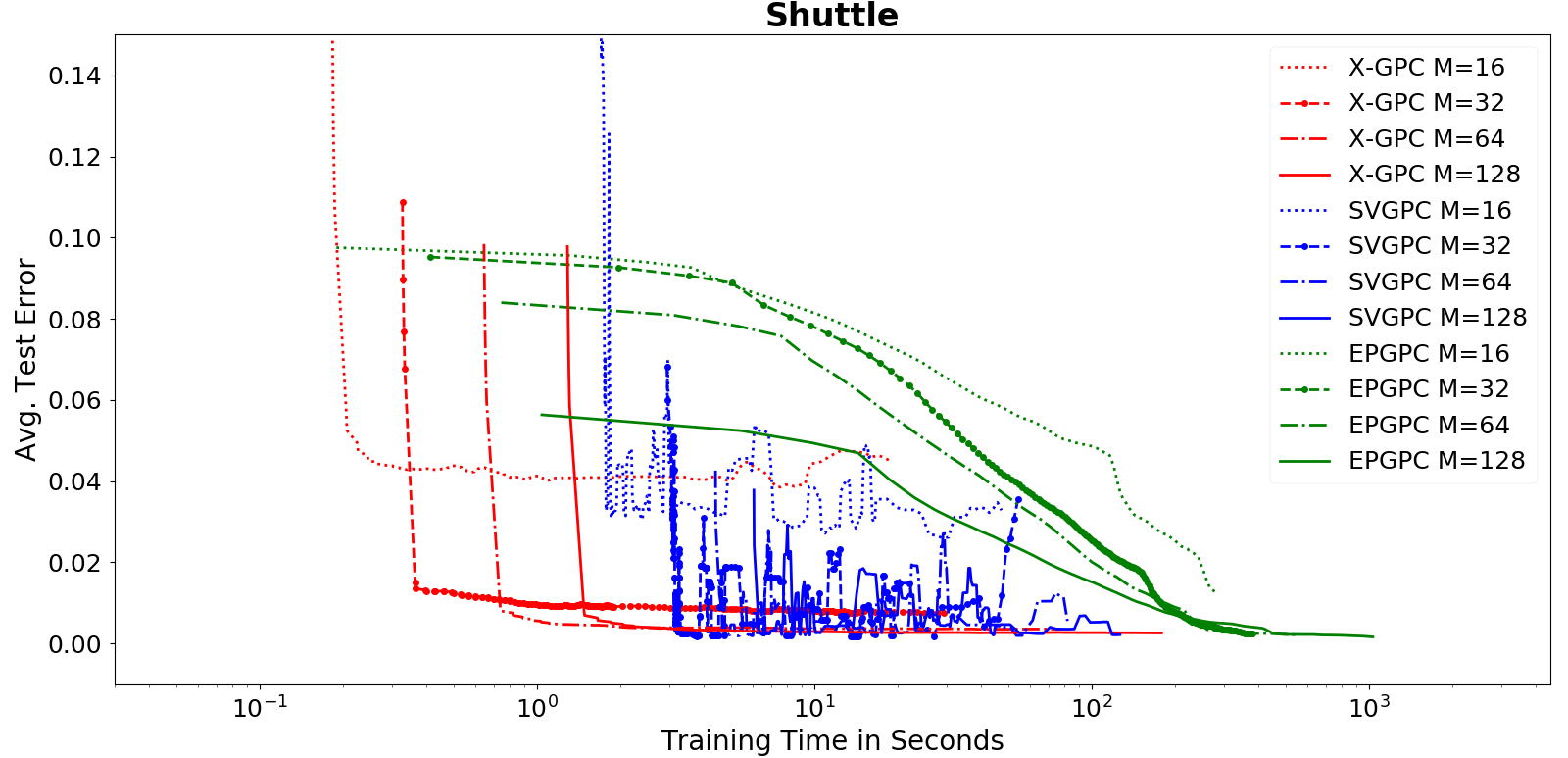}
\caption{Prediction error as function of training time (on a $\log_{10}$ scale) for the Shuttle dataset. Different numbers of inducing points are considered, $M = 16, 32, 64, 128$. \textsc{x-gpc} (proposed) converges the fastest in all settings of different numbers of inducing points. Using only 32 inducing points is enought for obtaining allmost optimal prediction performance for all methods, but \textsc{svgpc} becomes instable in settings of less than 128 inducing points.}
\label{fig:inducing_points}
\end{figure*}

We evaluate the prediction performance and run time of our method \textsc{x-gpc} and the competing methods \textsc{svgpc} and \textsc{epgpc}.
We experiment on a variety of different datasets and report the resulting prediction error, negative test log-likelihood and run time for each method in table~\ref{tab:performance}.

The experiments are conducted as follows.
For each dataset we perform a 10-fold cross-validation and for datasets with more than 1 million points, we limit the test set to 100,000 points. We report the average prediction error, the negative test log-likelihood \eqref{eq:test_distribution} and the run time along with one standard deviation.
For all datasets, we use 100 inducing points and a mini-batch size of 100 points.

For \textsc{x-gpc} we find that the following simple convergence criterion on the global parameters leads to good results:
 a sliding window average being smaller than a threshold of $10^{-4}$ .
Unfortunately, the original implementations of \textsc{svgpc} and \textsc{epgpc} do not include a convergence criterion. 
We find that the trajectories of the global parameters of \textsc{svgpc} tend to be noisy, and using a convergence criterion on the global parameters often leads to poor results.
To have a fair comparison, we therefore monitor the convergence of the prediction performance on a hold-out set and use a sliding window average of size 5 and threshold $10^{-3}$ as convergence criterion for all methods.

We observe that \textsc{x-gpc} is about one to two orders of magnitude faster than \textsc{svgpc} and \textsc{epgpc} on most datasets. Only on the dataset wXa, \textsc{epgpc} is slightly faster than \textsc{x-gpc}. 
The prediction error is similar for all methods but \textsc{x-gpc} outperforms the competitors in terms of the test log-likelihood on most datasets (aXa, Bank Marketing, Click Prediction, Cod RNA, Diabetes, Electricity, German, Higgs, Mnist, SUSY). This means that the confidence levels in the predictions are better calibrated for \textsc{x-gpc}, i.e. when predicting a wrong label \textsc{svgpc} and \textsc{epgpc} tend to be more confident than \textsc{x-gpc}.

\paragraph{Performance as a function of time}
Since all considered methods are based on an optimization schemes, there is a trade-off between the run time of the algorithm and the prediction performance. 
We make this trade-off transparent by
plotting the prediction performance as function of time on each dataset. For each method we monitor on a 10-fold cross-validation the average negative test log-likelihood and prediction error on a hold-out test set as a function of time.


The results are displayed in Fig.~\ref{fig:timeexperiment} for three selected datasets, while the results for the remaining datasets are deferred to appendix \ref{sec:appendix_more_plots}.
For all datasets we observe that after a few iterations \textsc{x-gpc} is already close to the optimum due to its efficient closed form natural gradient updates. Both the prediction error and test log-likelihood converge around one to two orders of magnitude faster for \textsc{x-gpc} than for \textsc{svgpc} and \textsc{epgpc}. Moreover, the performance curves tend to be noisier for \textsc{svgpc} than for \textsc{x-gpc} and \textsc{epgpc}.
For the datasets HIGGS and IJCNN, \textsc{epgpc} lead to slightly better final prediction performance, but with the cost of a runtime being up to 4 orders of magnitude slower than \textsc{x-gpc} (approx. 28 hours vs. 9 and 435 seconds, respectively).


All three methods are implemented in different programming frameworks: \textsc{x-gpc} in Julia, \textsc{svgpc} in TensorFlow and \textsc{epgpc} in R leading to different efficient implementations. However, we find that the main speed-up of our method is due to the efficient natural gradient updates and only marginally related to the usage of a different programming language. To check this we implemented \textsc{epgpc} also in Julia and obtained similar runtimes. Since \textsc{svgpc} is part of the highly optimized GPflow package we only used the original implementation. 

\subsection{Inducing points}
\label{sec:exp_inducing}
We examine the effect of different numbers of inducing points on the prediction performance and run time. For all methods we compare different numbers of inducing points: $M=16,32,64,128$. For each setting, we perform a 10-fold cross validation on the Shuttle dataset and plot the mean prediction error as function of time.
The results are displayed in Fig.~\ref{fig:inducing_points}.
We observe that the higher the number of inducing points, the better the prediction performance, but the longer the run time. Throughout all settings of inducing points our method is consistently faster of around one to two orders of magnitude than the competitors. On the Shuttle dataset using only $M=32$ inducing points is enough and can only be marginally improved by using more inducing point for all methods. However, the performance curves of \textsc{svgpc} are instable when using less than 128 inducing points.

\section{Conclusions} \label{sec:conclusion}
We proposed an efficient Gaussian process classification method that builds on P\'olya-Gamma data augmentation and inducing points.
The experimental evaluations shows that our method is up to two orders of magnitude faster than the state-of-the-art approach while being competitive in terms of prediction performance.
Speed improvements are due to the P\'olya-Gamma data augmentation approach that enables efficient second order optimization.

The presented work
shows how data augmentation can speed up variational approximation of GPs.  Our analysis may pave the way for using data augmentation to derive efficient stochastic variational algorithms also for variational Bayesian models other than GPs. Furthermore, future work may aim at extending the approach to multi-class and multi-label classification.


\subsubsection*{Acknowledgements}
We thank Stephan Mandt, James Hensman and Scott W. Linderman for fruitful discussions.
This work was partly funded by the German
Research Foundation (DFG) awards KL 2698/2-1 and GRK1589/2 and
the by the Federal Ministry of Science and Education (BMBF) awards 031L0023A, 01IS18051A.

\newpage
\fontsize{9.0pt}{10.0pt}
\selectfont
\bibliography{bib}
\bibliographystyle{aaai}

\newpage
\onecolumn
\appendix
\section{Appendix}


\subsection{Additional performance plots}
\label{sec:appendix_more_plots}
We show all time vs. prediction performance plots for the datasets presented in table~\ref{tab:performance} in section section~\ref{sec:exp_performance} which could not be included in the main paper due to space limitations.

\begin{figure}[h!]
\centering
\includegraphics[width=\textwidth]{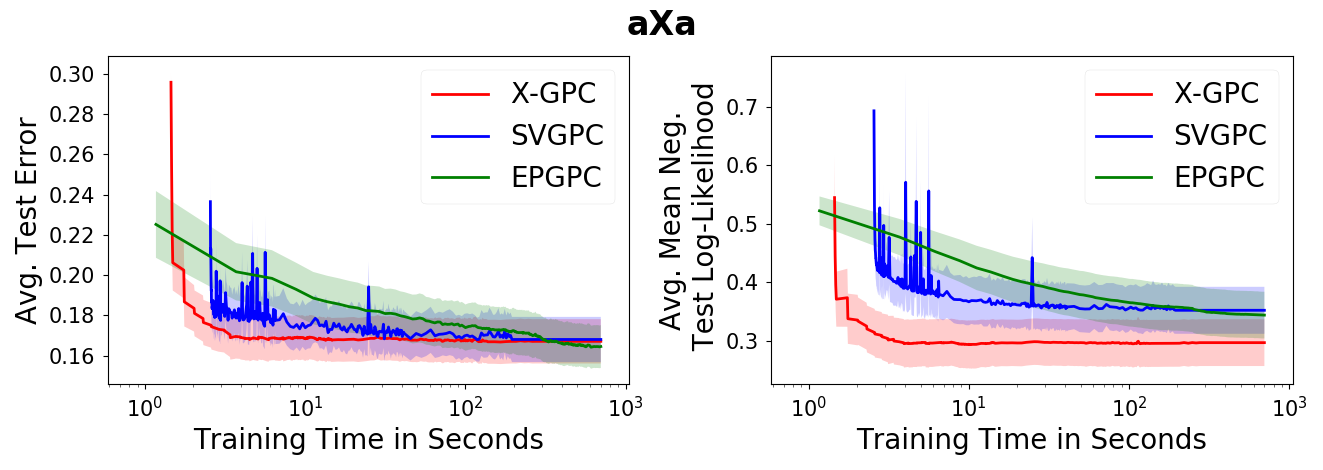}
\includegraphics[width=\textwidth]{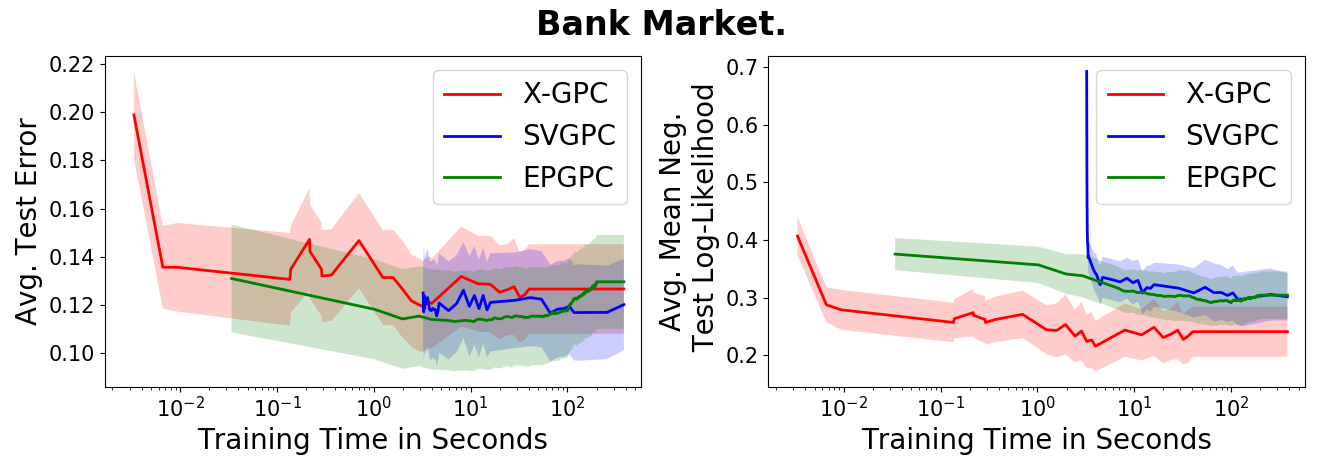}
\includegraphics[width=\textwidth]{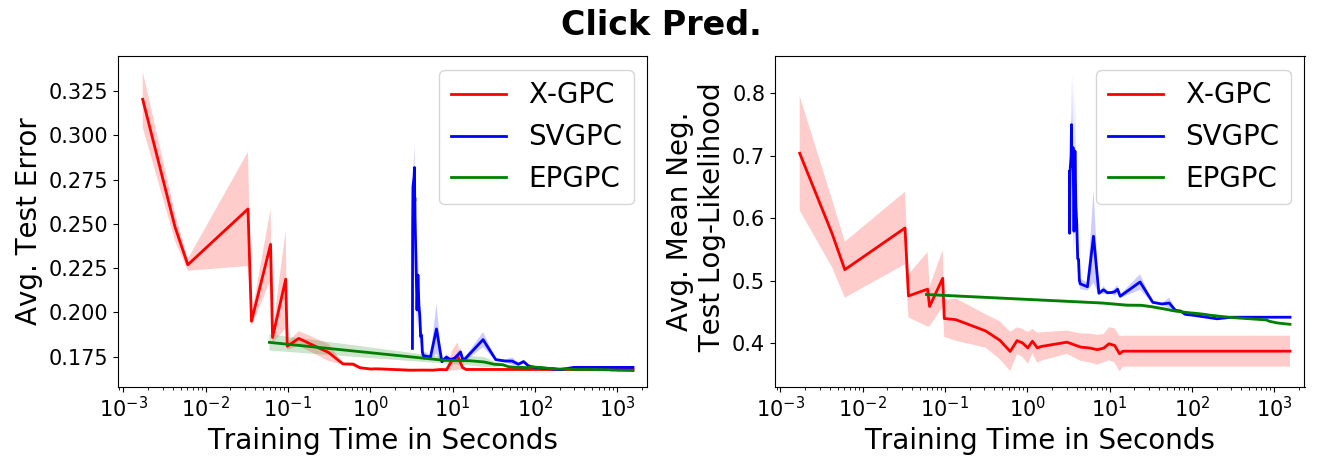}
\caption{Average negative test log-likelihood and average test prediction error as function of training time measured in seconds (on a $\log_{10}$ scale).}
\label{fig:performance_appendix1}
\end{figure}
\begin{figure}[h!]
\centering
\includegraphics[width=\textwidth]{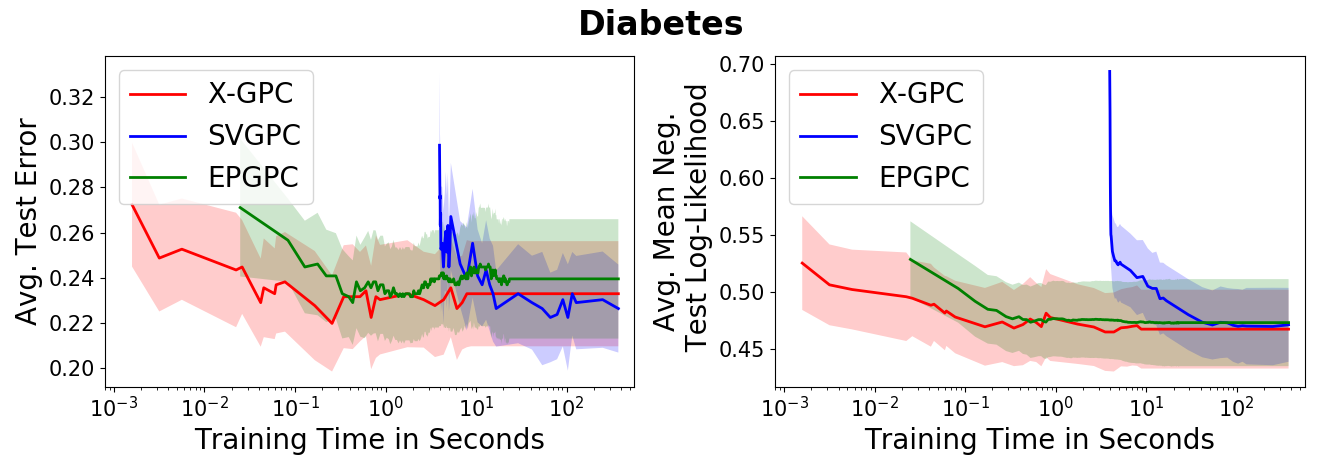}
\includegraphics[width=\textwidth]{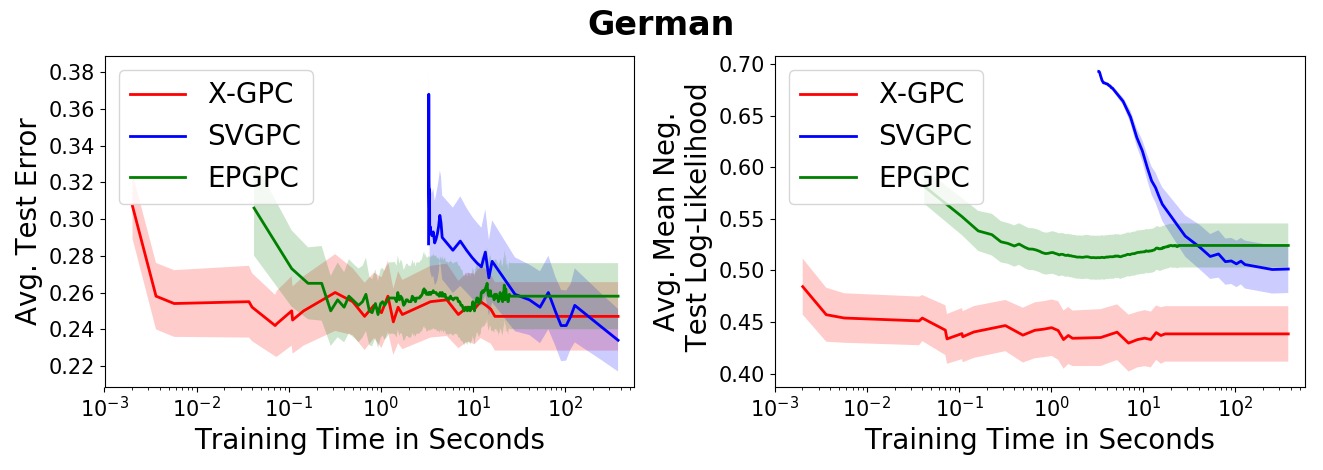}
\includegraphics[width=\textwidth]{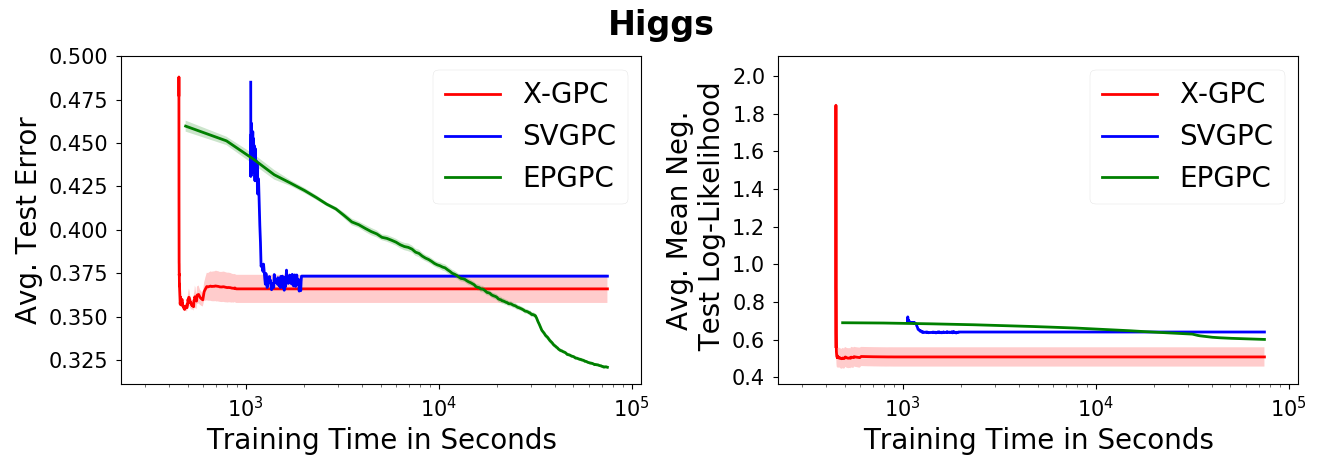}

\caption{Average negative test log-likelihood and average test prediction error as function of training time measured in seconds (on a $\log_{10}$ scale). For the dataset Higgs, \textsc{epgpc} exceeded the time budget of $10^5$ seconds ($\approx$ 28 h).}
\label{fig:performance_appendix2}
\end{figure}

\newpage

\begin{figure}[t]
\centering
\includegraphics[width=\textwidth]{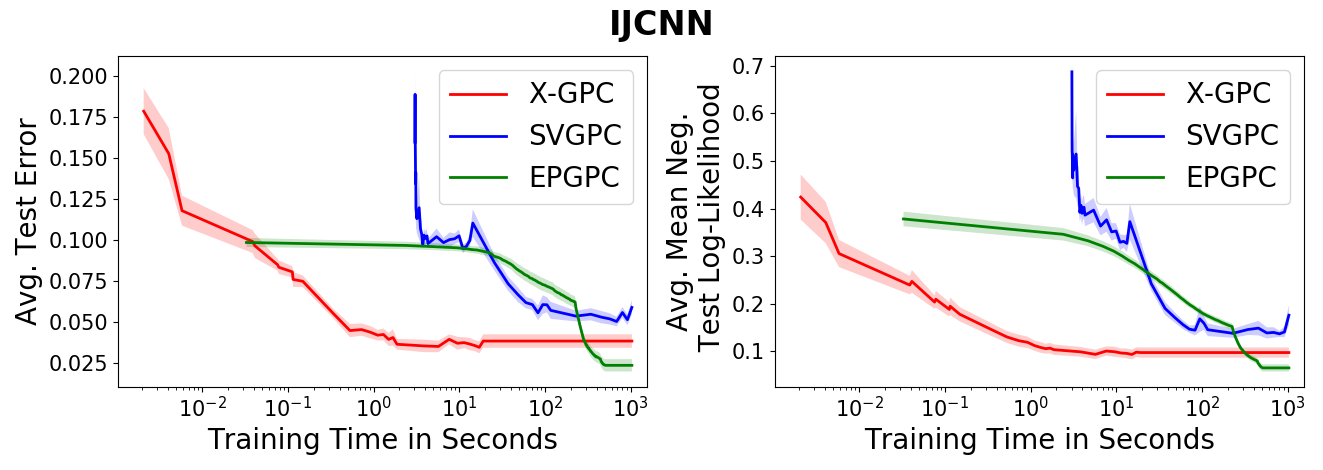}
\includegraphics[width=\textwidth]{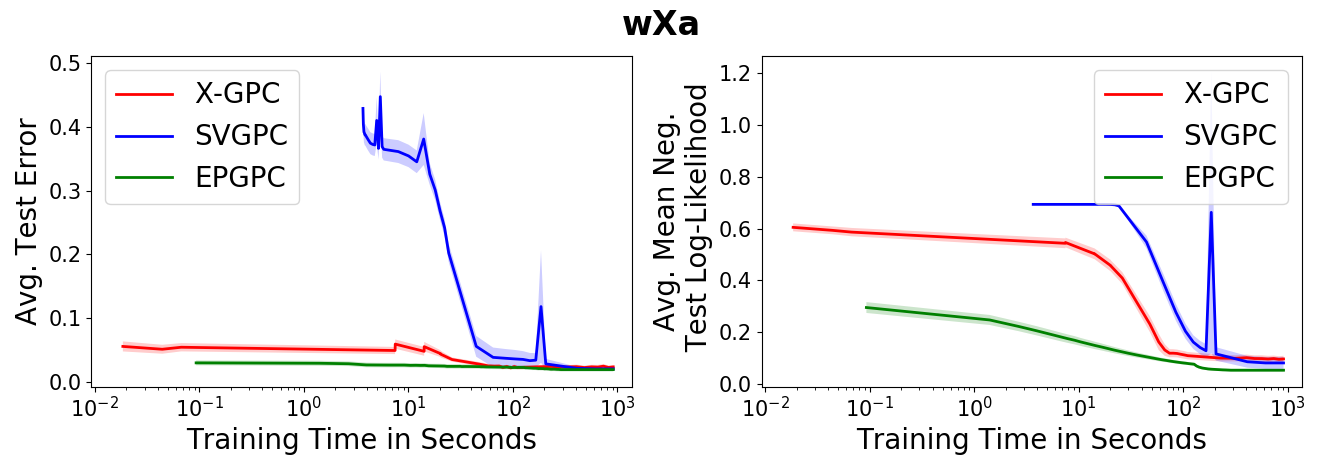}
\caption{Average negative test log-likelihood and average test prediction error as function of training time measured in seconds (on a $\log_{10}$ scale).}
\label{fig:performance_appendix2}
\end{figure}

\hspace{16cm}
\subsection{Variational bound}
\label{sec:appendix_variational_bound}
We provide details of the derivation of the variational bound \eqref{eq:variational_bound} which is defined as
\begin{align*}
\mathcal{L}(\boldc, \bmu, \Sigma) 
&= \EE_{p(\boldf|\boldu) q(\boldu) q(\bomega)}[\log p(\boldy | \bomega, \boldf)] - \mathrm{KL}\left( q(\boldu, \bomega) || p(\boldu, \bomega) \right),
\end{align*}
and the family of variational distributions
\begin{align*}
q(\boldu,\bomega) = q(\boldu)\prod_i q(\omega_i)
    = \mathcal{N}(\boldu|\bmu,\Sigma) \prod_i \mathrm{PG}(\omega_i|1,c_i).
\end{align*}
Considering the likelihood term we obtain
\begin{align*}
\EE_{p(f|u)}[\log p(\boldy | \bomega, f)]
    &\uptoconst \frac{1}{2}\expec{p(f|u)}{\boldsymbol{y}^\top\boldsymbol{f} - \boldsymbol{f}^\top\Omega\boldsymbol{f}}\\
	&= \frac{1}{2}\left(\boldy^\top \Knm \invKmm \boldu - \trace(\Omega\Ktilde) - \boldu^\top\invKmm K_{mn} \Omega \Knm \invKmm \boldu \right). \nonumber
\end{align*}
Computing the expectations w.r.t. to variational distributions gives
\begin{align*}
&\hspace{-1cm}\EE_{p(\boldf|\boldu) q(\boldu) q(\bomega)}[\log p(\boldy | \bomega, \boldf)]\\
\hspace{1.5cm}&\uptoconst \frac{1}{2}\EE_{q(\boldu) q(\bomega)}\left[\boldy^\top \Knm \invKmm \boldu - \trace(\Omega\Ktilde) - \boldu^\top\invKmm K_{mn} \Omega \Knm \invKmm \boldu \right]\\
&= \frac{1}{2}\EE_{q(\boldu)}\left[\boldy^\top \Knm \invKmm \boldu - \trace(\Theta\Ktilde) - \boldu^\top\invKmm K_{mn} \Theta \Knm \invKmm \boldu \right]\\
&=\frac{1}{2}\left[\boldy^\top\Knm\invKmm\bmu-\trace(\Theta\Ktilde)-\trace(\invKmm K_{mn}\Theta \Knm\invKmm\Sigma) - \bmu^\top\invKmm K_{mn}\Theta \Knm\invKmm\bmu\right]\\
&= \frac{1}{2}\sum_i\left( y_i\bkappa_i\bmu - \theta_i\Ktilde_{ii}-\theta_i\bkappa_i\Sigma \bkappa_i^\top -\theta_i\bmu^\top\bkappa_i^\top\bkappa_i\bmu \right),
\end{align*}
where $\theta_i = \EE_{p(\omega_i)}[\omega_i] = \frac{1}{2c_i}\tanh\left(\frac{c_i}{2}\right)$, $\Theta = \text{diag}(\boldsymbol{\theta})$ and $\bkappa_i = K_{im}\invKmm$. 

The Kullback-Leibler divergence between the Gaussian distributions $q(\boldu)$ and $p(\boldu)$ is easily computed
\begin{align*}
    \text{KL}(q(\boldu))||p(\boldu)) &\uptoconst \frac{1}{2}\left(\trace\left(\invKmm \Sigma\right)+\bmu^\top\invKmm \bmu -\log |\Sigma|+\log |K_{mm}|\right).
\end{align*}

The Kullback-Leibler divergence regarding the P\'olya-Gamma also can be computed in closed-form.
Have $q(\omega_i) = \text{cosh}\left(\frac{c_i}{2}\right)\exp\left(-\frac{c_i^2}{2}\omega_i\right)\mathrm{PG}(\omega_i|1,0)$ and $p(\omega_i) = \mathrm{PG}(\omega_i|1,0)$ we obtain
\begin{align*}
\text{KL}(q(\bomega))||p(\bomega)) &= \expec{q(\bomega)}{\log q(\bomega) - \log p(\bomega)} \\
&= \sum_i \left( \expec{q(\omega_i)}{\log\left( \text{cosh}\left(\frac{c_i}{2}\right)\exp\left(-\frac{c_i^2}{2}\omega_i\right)\mathrm{PG}(\omega_i|1,0) \right)}   -  \expec{q(\omega_i)}{\log \mathrm{PG}(\omega_i|1,0)}\right)\\
&= \sum_i \left( \log \text{cosh}\frac{c_i}{2} -\frac{c_i}{4}\text{tanh}\left(\frac{c_i}{2}\right) + \expec{q(\omega_i)}{\log \mathrm{PG}(\omega_i|1,0)} - \expec{q(\omega_i)}{\log \mathrm{PG}(\omega_i|1,0)}\right)\\
&= \sum_i \left( \log \text{cosh}\frac{c_i}{2} -\frac{c_i}{4}\text{tanh}\left(\frac{c_i}{2}\right)\right).
\end{align*}
Remarkably, the intractable expectations cancel out which would not have been the case if we assumed $\mathrm{PG}(\omega_i|b_i,c_i)$ as variational family. In section \ref{sec:var_approximation} we have shown that the restricted family $b_i=1$ contains the optimal distribution.

Summing all terms results in the final lower bound
\begin{align*}
\mathcal{L}(\boldc, \bmu, \Sigma) 
&\,\uptoconst \frac{1}{2}\bigg(\log|\Sigma|-\log|K_{mm}|)- \trace(\invKmm\Sigma)\nonumber -\bmu^\top\invKmm\bmu+\\ 
&\quad\,\,\sum_i \Big\{y_i\bkappa_i\bmu - \theta_i\Ktilde_{ii}-\theta_i\bkappa_i\Sigma\bkappa_i^\top
-\theta_i\bmu^\top\bkappa_i^\top\bkappa_i\bmu+ c_i^2\theta_i - 2 \log\text{cosh}\frac{c_i}{2}\Big\}\bigg).
\end{align*}

\subsection{Variational updates}
\label{sec:appendix_updates}
\paragraph{Local parameters}
The derivative of the variational bound \eqref{eq:variational_bound} w.r.t. the local parameter $c_i$ is
\begin{align*}
\frac{\dd\mathcal L}{\dd c_i} &= \frac{1}{2} \frac{\dd}{\dd c_i} \left\{\theta_i\left(-\Ktilde_{ii}-\bkappa_i\Sigma\bkappa_i^\top
-\bmu^\top\bkappa_i^\top\bkappa_i\bmu+ c_i^2\right) - 2 \log\text{cosh}\frac{c_i}{2} \right\}\\
&= \frac{1}{2}\frac{\dd}{\dd c_i} \left\{\frac{1}{2c_i}\tanh\left(\frac{c_i}{2}\right)\left(-\Ktilde_{ii}-\bkappa_i\Sigma\bkappa_i^\top
-\bmu^\top\bkappa_i^\top\bkappa_i\bmu+ c_i^2\right) - 2 \log\text{cosh}\frac{c_i}{2} \right\}\\
&= \frac{\dd}{\dd c_i} \left\{\frac{1}{4c_i}\tanh \left(\frac{c_i}{2}\right)\left(\underbrace{-\Ktilde_{ii}-\bkappa_i\Sigma\bkappa_i^\top - \bmu^\top\bkappa_i^\top\bkappa_i\bmu}_{:=-A_i}\right) + \frac{c_i}{4}\tanh \left(\frac{c_i}{2}\right) - \log \cosh \frac{c_i}{2} \right\}\\
&= \left(\frac{A_i}{4c_i^2} -\frac{1}{4}\right)\tanh \left(\frac{c_i}{2}\right)- \frac{1}{2}\left(\frac{A_i}{4c_i} -\frac{c_i}{4}\right)\left(1-\tanh^2(\frac{c_i}{2})\right)\\
&= U(c_i)\left(\frac{c_i}{2}\left(1-\tanh^2(\frac{c_i}{2})\right) - \tanh \left(\frac{c_i}{2}\right)\right),
\end{align*}
where $U(c_i)=\frac{\Sigma_{ii} + \mu_i^2}{4c_i^2} -\frac{1}{4}$.

The gradient equals zero in two case. First, in the case $U(c_i)=0$ which leads to\footnote{We omit the negative solution since $\text{PG}(b,c) = \text{PG}(b,-c)$.}
$$c_i = \sqrt{\Ktilde_{ii}+\bkappa_i\Sigma\bkappa_i^\top + \bmu^\top\bkappa_i^\top\bkappa_i\bmu},$$
which is always valid since $\bkappa$, $\Sigma$ and $\Ktilde$ are definite positive matrices.
The second consists of the right hand side of the product being zero which leads to
$c_i=0$.
The second derivative reveals that the first case always corresponds to a maximum and the second case to a minimum. 

\paragraph{Global parameters}
We first compute the Euclidean gradients of the variational bound \eqref{eq:variational_bound} w.r.t. the global parameters $\bmu$ and $\Sigma$. We obtain
\begin{align}
\begin{split} \label{eq:eucl_grad_Sigma}
\frac{d\mathcal{L}}{d\bmu} &= \frac{1}{2}\frac{d}{d\bmu}\left(-\bmu^\top\invKmm\bmu + \boldy^\top\bkappa\bmu -\bmu^\top\bkappa^\top\Theta\bkappa\bmu \right)\\
&= \frac{1}{2}\left(-2\invKmm\bmu+\bkappa^\top\boldy-2\bkappa^\top\Theta\bkappa\bmu\right)\\
&= -\left(\invKmm + \bkappa^\top\Theta\bkappa\right)\bmu+\frac{1}{2}\bkappa^\top\boldy,
\end{split}
\end{align}
and
\begin{align}
\begin{split} \label{eq:eucl_grad_mu}
\frac{d\mathcal{L}}{d\Sigma} &= \frac{1}{2}\frac{d}{d\Sigma}\left(\log \,\vert\,\Sigma\,\vert\,-\trace(\invKmm\Sigma)-\trace(\bkappa^\top\Theta\bkappa\Sigma)\right)\\
&= \frac{1}{2}\left(\Sigma^{-1}-\invKmm -\bkappa^\top\Theta\bkappa\right).
\end{split}
\end{align}

We now compute the natural gradients w.r.t. natural parameterization of the variational Gaussian distribution, i.e the parameters $\boldeta_1:=\Sigma^{-1} \bmu$ and $\eta_2 = -\frac{1}{2}\Sigma^{-1}$.
For a Gaussian distribution, properties of the Fisher information matrix expose the simplification that the natural gradient w.r.t. the natural parameters can be expressed in terms of the Euclidean gradient w.r.t. the mean and covariance parameters. It holds that  \begin{align}
   \widetilde\nabla_{(\boldeta_1, \eta_2)} {\cal L}(\eta) = \big(\nabla_\mu{\cal L}(\eta) - 2\nabla_\Sigma{\cal L}(\eta)\mu,\; \nabla_\Sigma{\cal L}(\eta)\big), \label{eq:nat_grad_Gaussian}
\end{align}
where $\widetilde\nabla$ denotes the natural gradient and $\nabla$ the Euclidean gradient.
Substituting the Euclidean gradients \eqref{eq:eucl_grad_mu} and \eqref{eq:eucl_grad_Sigma} in to equation \eqref{eq:nat_grad_Gaussian} we obtain the natural gradients
\begin{align*}
    \widetilde\nabla_{\eta_2} {\cal L} &= \frac{1}{2}\left(-2\eta_2 -\invKmm -\bkappa^\top\Theta\bkappa\right)\\
    &=-\eta_2 - \frac{1}{2}\left(\invKmm + \bkappa^\top\Theta\bkappa\right)
\end{align*}    
and
\begin{align*}
    \widetilde\nabla_{\boldeta_1} {\cal L} &= -\left(\invKmm + \bkappa^\top\Theta\bkappa\right)(-\frac{1}{2}\eta_2^{-1}\boldeta_1) +\frac{1}{2}\bkappa^\top\boldy - 2 \left( -\eta_2 - \frac{1}{2}\left(\invKmm + \bkappa^\top\Theta\bkappa\right)\right)(-\frac{1}{2}\eta_2^{-1}\boldeta_1)\\
    &= \frac{1}{2}\bkappa^\top\boldy - \boldeta_1.
\end{align*}

\subsection{Natural gradient and coordinate ascent updates}
\label{sec:appendix_natural_grad}
If the full conditional distributions and the corresponding variational distribution belong to the same exponential family it is known in variational inference that
``we can compute the natural gradient by computing the coordinate updates in parallel and subtracting the current setting of the parameter'' \cite{JMLR:v14:hoffman13a}.
In our setting it is not clear if this relation holds since we do not consider the classic ELBO but a lower bound on it due to \eqref{eq:ineq_likelihood}. Interestingly, the lower bound \eqref{eq:ineq_likelihood} does not break this property and our natural gradient updates correspond to coordinate ascent updates as we show in the following.
Setting the Euclidean gradients and \eqref{eq:eucl_grad_Sigma} to zero and using the natural parameterization gives
\begin{align}
    \eta_2 = -\frac{1}{2}\Sigma^{-1} = -\frac{1}{2}\left(\invKmm +\bkappa^\top\Theta\bkappa\right).\label{eq:cavi_eta2}
\end{align}
Setting \eqref{eq:eucl_grad_mu} to zero yields
\begin{align*}
    \bmu &= \frac{1}{2}\left(\invKmm+\bkappa^\top\Theta\bkappa\right)^{-1}\bkappa^\top\boldy.
\end{align*}
Substituting the update from above \eqref{eq:cavi_eta2} and using natural parameterization results in
\begin{align*}
\boldeta_1 = \frac{1}{2}\bkappa^\top\boldy.
\end{align*}
This shows that using learning rate one in our natural gradient ascent scheme corresponds to employing coordinate ascent updates in the Euclidean parameter space.


\subsection{Variational bound by Gibbs and MacKay}
\label{sec:appendix_mackay}
When using the full GP representation in our model and not the sparse approximation, the bound in our model is equal to the bound used by \citet{Gibbs_MacKay97b}. We provide a proof in the following.

Applying our variational inference approach to the joint distribution \eqref{eq:pg_joint} gives the variational bound
\begin{equation*}
\begin{split}
    \log p(\boldy \vert\,\boldf) &\ge \EE_{q(\omega)}\left[\log p(\boldy\vert\,\boldf,\bomega)\right] - \mathrm{KL}(q(\bomega)\vert p(\bomega))\\
    &= \EE_{q(\omega)}\left[\frac{1}{2}\boldy^\top\boldf - \frac{1}{2}\boldf^\top \Omega\boldf\right] -  n\log(2) - \mathrm{KL}(q(\bomega)\vert p(\bomega)) \\
    &=  \frac{1}{2}\boldy^\top\boldf - \frac{1}{2}\boldf^\top \Theta\boldf - n\log(2) + \sum_{i=1}^n\left( \frac{c_i^2}{2}\theta_i - \log\cosh(c_i/2)\right).
\end{split}
\end{equation*}

\cite{Gibbs_MacKay97b} employ the following inequality on logit link 
\begin{align*}
    \sigma(z) \geq & \sigma(c)\exp\left(\frac{z-c}{2} - \frac{\sigma(c) - 1/2}{2c}(z^2-c^2)\right).
\end{align*}
Using this bound in the setting of GP classification yields the following lower bound on the log-likelihood,
\begin{equation*}
\begin{split}
    \log p(\boldy \vert\,\boldf) = & \sum_{i=1}^n \log \sigma(y_i f_i)\\
    \geq & \sum_{i=1}^n\left(\log\sigma(c_i) + \frac{y_if_i-c_i}{2} - \frac{\sigma(c_i) - 1/2}{2c_i}((y_i f_i)^2-c_i^2)\right)\\
    = & \sum_{i=1}^n\left(- \log \cosh(c_i/2) - \log(2) + \frac{y_if_i}{2} - \frac{\sigma(c_i) - 1/2}{2c_i}(f_i^2-c_i^2)\right) \\
    = & \sum_{i=1}^n\left(- \log \cosh(c_i/2) - \log(2) + \frac{y_if_i}{2} - \frac{1}{4c_i}\tanh(c_i/2)(f_i^2-c_i^2)\right) \\
    = & \sum_{i=1}^n\left(- \log \cosh(c_i/2) - \log(2) + \frac{y_if_i}{2} - \frac{1}{2}\theta_i(f_i^2-c_i^2)\right) \\
    = & \frac{1}{2}\boldy^\top\boldf - \frac{1}{2}\boldf^\top \Theta\boldf - n\log(2) + \sum_{i=1}^n\left( \frac{c_i^2}{2}\theta_i - \log\cosh(c_i/2)\right),
\end{split}
\end{equation*}
where we made use of the fact that $\sigma(x) - 1/2 = \tanh(x/2)/2$. 
This concludes the proof.

\end{document}